\documentclass{svproc}
\usepackage{url}

\usepackage{graphicx}
\usepackage{subfig}
\usepackage[bottom]{footmisc}
\usepackage{array}
\newcommand{\PreserveBackslash}[1]{\let\temp=\\#1\let\\=\temp}
\newcolumntype{C}[1]{>{\PreserveBackslash\centering}m{#1}}
\newcolumntype{R}[1]{>{\PreserveBackslash\raggedleft}m{#1}}
\newcolumntype{L}[1]{>{\PreserveBackslash\raggedright}m{#1}}
\usepackage{amsmath}
\begin{document}
\mainmatter              
\title{Real-Time Face Recognition System for Remote Employee Tracking}
\titlerunning{Real-Time Face Recognition System for Remote Employee Tracking}  
%
\author{Mohammad Sabik Irbaz \and MD Abdullah Al Nasim \and Refat E Ferdous}
\authorrunning{Irbaz et. al.} 
%
\institute{Machine Learning Team, Pioneer Alpha Ltd\\
\email{sabikirbaz@iut-dhaka.edu}\\
\email{nasim.abdullah@ieee.org} \\
\email{refat.amarischool@gmail.com} \\
}

\maketitle              

\begin{abstract}
During the COVID-19 pandemic, most of the human-to-human interactions have been stopped. To mitigate the spread of deadly coronavirus, many offices took the initiative so that the employees can work from home. But, tracking the employees and finding out if they are really performing what they were supposed to turn out to be a serious challenge for all the companies and organizations who are facilitating "Work From Home". To deal with the challenge effectively, we came up with a solution to track the employees with face recognition. We have been testing this system experimentally for our office. To train the face recognition module, we used FaceNet with KNN using the Labeled Faces in the Wild (LFW) dataset and achieved 97.8\% accuracy. We integrated the trained model into our central system, where the employees log their time. In this paper, we discuss in brief the system we have been experimenting with and the pros and cons of the system.
\end{abstract}

\section{Introduction}
Face recognition is an innovative identifying system that works on recognizing the face of a potential individual after comparing it with a source image stored in a database by a vision system. This technology extracts features from an input image and recognizes them by detecting the specific and particular information about an individual’s face. We can see many applications of this, for example, access and security, criminal identification, online payments, design of a smart home, etc. In the face recognition and detection process, no human cooperation is needed, which makes it the best method for person identification \cite{marcialis2003fusion}. There are various biometric authentication systems available, face recognition is one of them. In biometric-based strategies, one’s physiological and behavioural attributes have been analyzed with a specified end goal to determine their uniformity \cite{kak2018review}. Hence compared to other biometric-based systems, face recognition has been considered as one of the most significant biometric authentication techniques among the most constitutive applications. \\

\noindent Starting from 2020, during the COVID-19 pandemic, many offices took the initiative so that the employees can work from home. But, tracking the employees and finding out if they are really there to work had been a serious challenge for the administration. Existing time tracker solutions can be easily fooled, which we have already seen in our company. To deal with the challenge and build a system that cannot be fooled, we came up with a solution to track the employees with face recognition. We have been testing this system internally. To train the face recognition module, we used FaceNet with KNN using the Labeled Faces in the Wild (LFW) dataset \cite{learned2016labeled}.\\

\noindent We organized our paper as follows: In section II, we presented the existing literature methods that helped us to reach our goal. In section III, we gave a description of the dataset we used in our model. Section IV provides a summary of our methodology. Experimental evaluations have been discussed in section V. Finally, in section VI, we give a brief discussion on future work and conclusion.\\

\section{Related Works}

\noindent In a review till 2006 on face recognition \cite{tolba2006face}\cite{schreiber1991facenet}, we can find various databases for face recognition. This is the start of face recognition revolution. We find more interesting works in 2011 by Deniz et. al. \cite{deniz2011face}. As HOG operates on local cells, it is invariant to geometric transformations and photometric transformations, except for object orientation. This experiment has used single angle orientation to allow more differentiation between patterns. Overlapping in HOG significantly improves the performance of detection and identification which would have been otherwise quite difficult in presence of poor lighting conditions. To remove redundancy in the data and avoid over fitting, they proposed to use, dimensionality reduction in the HOG representation. To provide robustness to facial feature detection, they propose uniform sampling of the HOG features. \\

\noindent Fast forward to very recent works we can find Prasad et. al. \cite{prasad2020deep} using Deep Learning based representations. They input raw data in convolve filters in different levels that automatically detect underlying high levels from labelled or unlabeled data. Face verification approaches classified into three sets. First one is used to extract face feature vectors and process them by classifiers. Second one directly enhances the proof loss for matching and non-matching pairs. Third one combined identification and verification constraints to improve the deep face model. For identifying faces two approaches are explained here. First one is effective Convolutional Neural Network designs for biometric face recognition. Second one is, explanation of a face plan which is based on two models (VGG-Face and Lightened CNN). This study has discovered that Deep Learning based face recognition is more robust to misalignment facial images. \\

\section{Literature Review}

\subsection{FaceNet} 
FaceNet model was developed by Google researchers Schroff et al., which can reduce the difficulties in face detection and verification process and achieve the desired result \cite{jose2019face}. The FaceNet algorithm works by taking an input of an image of a person’s face and converting the high dimensional vectors of that image into a 128 low dimensional Euclidean Space. This conversion process is also known as embedding. The FaceNet model uses the Deep Convolutional Networks, which makes the most effective use of its embedding, compared to using intermediate bottleneck layers as a test of previous deep learning approaches \cite{william2019face}. This method is known as one-shot learning. In this way, triplet loss has been trained over the produced FaceNet model, which can achieve a result of uniformity and distinction over the given collections of images. After collecting the result of similarities between the faces we can consider the FaceNet embedding as feature vectors. After creating the vector space, operations like face recognition, classification, verification, and clustering could be implemented more spontaneously. Additionally, FaceNet is able to train any difficult learning system of any single model that illustrates an entire goal system by collecting all of the factors at the same time and this is the most significant part of the FaceNet model. \\

\subsection{Triplet Loss}
In Deep Neural Networks, the triplet-based loss function used to learn the mapping is an adaption of Kilian Weinberger’s Large Margin Nearest Neighbor (LMNN) classifier \cite{schroff2015facenet}. Triplet loss is one of the best ways to learn good 128-dimensional embedding for each individual face. \cite{weinberger2009distance}  As it is an equivalent of the loss function, a comparison between the baseline input with positive input and negative input has been observed here. That’s why to compare to other loss functions, triplet loss is more compatible with face verification. \\
\noindent The triplet loss function has been categorized into three main parts. They are:
\begin{itemize}
    \item Anchor
    \item Positive
    \item Negative
\end{itemize}

\noindent Here anchor is the starting input point, which is used for comparison. Positive denote similar identity like the anchor and negative denotes the different identity from the anchor. The distance between the anchor and a chosen image will be minimized if that image is positive, which denotes that the compared two images have a similar identity. In contrast, the distance will be maximized for different images.\\

\subsection{Triplet Selection}
Triplet selection is a core part of improving the performance of triplet loss. So, it is essential to select all hard triplets. Training with randomly hard triplets almost does not converge whereas training with the hardest triplets often leads to a bad local solution. There are two types of triplet loss methods. They are online and offline triplet mining methods. Offline triplet mining worked with a full pass on the training set to generate triplets. Since it needs to update regularly, this method is not efficient. So, we used online triplet mining.\\
First of all, we picked all the necessary triplets. Depending on the loss we classify the triplets in three categories: easy, semi-hard, hard. As easy triplets are with loss 0 or very small loss so, we intended not to take them.  Next, the semi-hard triplets are also far away from the anchor than the ideal model. In a hard batch, we select a set of hard anchor-positive pairs and then select the hardest negatives within the mini-batch. Here small mini-batches are used since they improve the process of converging. \\

\subsection{MTCNN}
MTCNN or multi cascade convolutional neural network is an improved neural network algorithm in the detection of faces and facial landmarks on images. The purpose of the proposed MTCNN is to form an avalanched structure and use it as material for multi-task knowledge to forecast the location of the face and it is marked in a coarse-to-fine method. And in its application, MTCNN is able to detect real-time with fairly high accuracy [13]. The MTCNN architecture is made up of three convolutional networks, which are connected in a cascade. The first one is, Proposal Network(P-Net), which main task is giving a boundary box for each face bounds. In this process, a large number of face detection and false detection has been taken place. The second one is, Refine Network(R-Net), which is slightly similar to P-Net. But R-Net includes more appropriate bounding boxes compare to P-Net. Thus, making a refinement of the result by eliminating most of the false detection and aggregate bounding boxes [12]. The last one is, Output Network (O-Net), which produces an output different from P-Net and R-Net. There are three types of output that O-Net produces. The output of the first layer is used for measuring face probabilities in the box. The second layer is used to give the boundary coordinates in the box. And the last layer is used for the coordinates of the five landmarks of the faces \cite{william2019face}. \\

\section{Dataset}

We used Labeled Faces in the wild (LFW) \cite{learned2016labeled} here to uphold the effectiveness of the face recognition process. It has been said that the LFW dataset is a standard for the face verification process which is mainly developed to perusing the unconstrained problem of face recognition. This dataset consists of around 13K images collected from the web. Also, each image has been labeled with the name of that specific person. There are four different sets of LFW images including the original and three different types of ‘aligned’ images. Among them, deep funneled images produce superior results for most face verification algorithms over the original and funneled images. Hence, the dataset we used here is the deep funneled version.\\ \\

\begin{figure}[h]
    \centering
    \includegraphics[width=\linewidth]{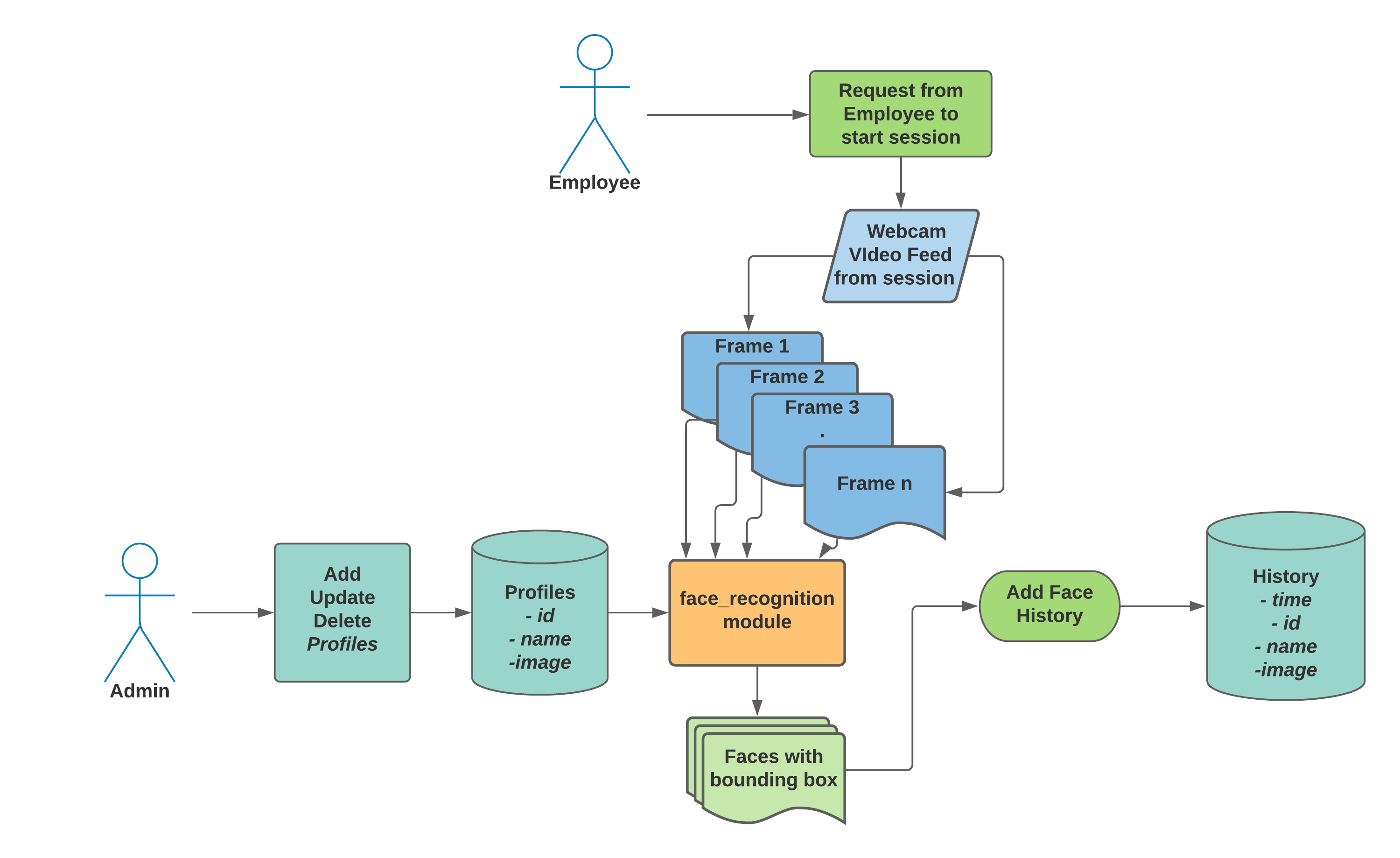}
    \caption{Full Pipeline overview of our methodology}
    \label{methodology}
\end{figure}

\section{Methodology} 
\noindent To track the presence of the employees, we initially put all the information and images of the employees in our central database. Then we used those images during the face recognition period. Later, we took the history and kept that in a separate relational database. So, in short, our system has three stages.\\

\subsection{Adding, Updating and Deleting Employee Details}
An admin can add new users when a new employee joins the company, update user details if anything changes related to any employee, and delete employee details when an employee leaves the company. While doing so, the face recognition system does not need to change anything. The recognition system can adapt with the user changes dynamically.\\

\subsection{Face Recognition}
\begin{figure}[h]
    \centering
    \includegraphics[width=\linewidth]{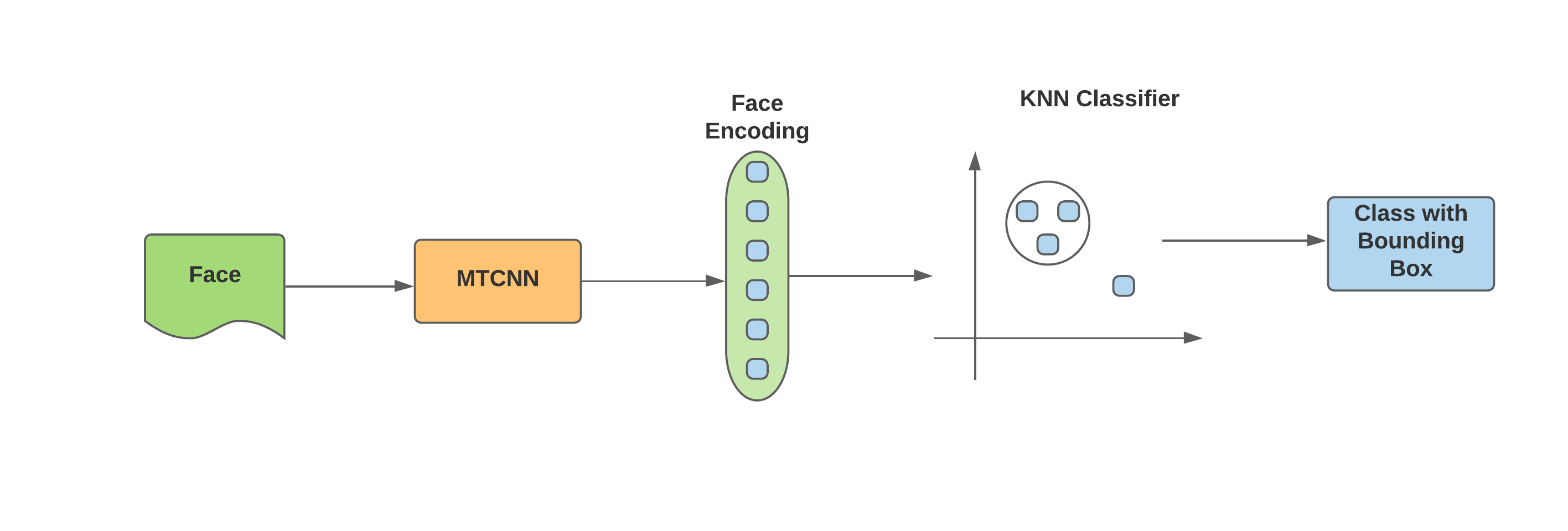}
    \caption{Face Recognition Module}
    \label{mod}
\end{figure}

We used the module in Figure \ref{mod}. We trained the MTCNN with triplet loss to train the module to detect face encodings from faces. After that, we compared the employee images using the KNN classifier. This returned us with the class i.e name of the person and bounding box identifying where the faces are. \\

\subsection{Tracking and Storing Face Recognition History} 
Every time any employee starts their work session. The video feed is taken from the webcam randomly from different portions of the daily sessions. If an employee who started his session is not found in video feeds consecutively then an automatic alert goes to the admin and the employee. All the face recognition history is stored as archive with employee details. Admins are not allowed to access the archive to keep the trasparency between employees.\\

\newpage

\section{Experiment Analysis}

\textbf{Experimental Setup} \\ \\
We did all our experiments using Google Colab which is a hosted Jupyter notebook service. We used it because Google Colab provides free GPU for 12 hours a day. In our experiments we used Numpy, Pandas, etc. for data processing and Tensor for training and testing. PyTorch \cite{paszke2017automatic} is  an open source machine learning framework. We chose 10\% randomly as validation set from the LFW dataset. We trained all the models for 10 epochs with Adam Optimizer \cite{kingma2014adam}. After taking a pair of two face images, we measured a squared L2 distance threshold, which is mainly used for classifying the similar and different images. The optimal threshold value we got here is 1.24. That means this threshold value can accurately classify all pairs of images. \\

\noindent\textbf{Result Analysis} \\ \\
Here, 128-dimensional float vector was used throughout the training season. This 128-byte dimensional vector tightly illustrate each face, which is most suitable for recognition and clustering. Here we took input images of size 220*220 pixels and trained our model through this. But still, it works good on 80*80 images, which is admissible.  Table 1 shows the size of training images and their corresponding accuracies.\\

\begin{table}[h]
\centering
\caption{Effect of training samples on performance}
\begin{tabular}{|c|c|}
\hline
\textbf{No. of Training Images} & \textbf{Accuracy} \\ \hline
70                           & 90.5              \\ \hline
700                          & 92.3              \\ \hline
7,000                         & 95.6              \\ \hline
Full Dataset                       & \textbf{97.8}              \\ \hline
\end{tabular}
\end{table}

\noindent Finally, face recognition task was completed by our pre-trained model based on FaceNet model. A number of facial encoding was generated by our trained model and by using Euclidean distance, which can show a comparison of facial coordinates for different samples. This results in a distance, which directly assembles to the measure of resemblance. This acquainted our model whether the face of input image matches with anyone or not. And that’s how we trained our model. For LFW dataset, we achieved a classification accuracy of 97.8\% . So, we can say that, in face recognition method, FaceNet with KNN shows up the accurate results with confident. 

\section{Qualitative Evaluation}
In Figure \ref{methodology}, we get a full overview of our working prototype. Our prototype takes the webcam video feed given by users which passes through the Flask \cite{grinberg2018flask} backend to the Face Recognition module. Later, we get the face encoding and bounding box from the module. We store detected faces in the database as history. We used this system experimentally in our company on remote employees. In short, they also acted as evaluators. Our backend can effectively extract faces and recognize them in a dynamic way. The evaluators were mainly interested in how well the software can perform rather than the user interface. The software was very much self-explanatory and most of our evaluators were expert enough. We asked them to look through the eyes of general people. They did not face any confusion while evaluating. Some evaluators complained that if they were looking somewhere else during random time video feed extraction, their faces were not detected. This is a limit of the dataset also which we plan to extend in the future. However, in most cases, even with blurry cameras our system could identify the faces. We did not find any case where our model predicted the wrong evaluator. We found out that even when our evaluators added 10 years old images, our system could find them. 

\section{Applications}
Even though we used this whole pipeline for remote employee tracking. This can be used in various other cases:
\begin{itemize}
    \item \textbf{Access and Security:} \\
    a) Instead of using passcode, mobile phones, laptops and other consumer electronics will be accessed via the owner's facial feature. Apple, Samsung, Xiaomi already installed FaceTech in their phones.\\
    b) In the near future, consumers will get into their cars, houses and other physical locations simply by looking at them. \\
    c) Innovative facial security could be helpful for any company or organization where sensitive data needs to keep tight control on who enters their facilities.\\ 
    \item \textbf{Payments:}\\
    a) In 2016, Mastercard launched a new selfie pay app called “Mastercard Identity” check. Customers open the app to confirm a payment using the camera.\\
    b) With FaceTech, customers wouldn’t even need their cards.\\
    c) In the future, we can do the same for online payments. \\
    d) Automatic face recognition to prohibit deduplication of identity to authentication of mobile payment. This is mainly used for face spoof attacks also known as biometric sensor presentation attacks, where a photo or video of an authorized person’s face could be used to gain access.\\
    \item \textbf{Criminal Identification:} FaceTech can be used to keep unauthorized people out of facilities.\\
    \item \textbf{Smart Home:} The design of smart homes or cities has become one of the things that many researchers have focused on. Especially people with special needs or patients.\\
    \item \textbf{Video Surveillance:} \\
    a) Surveillance used for protection, intelligent gathering, searching for drug offenders, CCTV control, power grid surveillance.\\
    b) CCTV cameras can be used to monitor any well-known criminals and authorities are notified if one is located. But this is quite challenging for light illumination, pose variation and facial expression.\\
    \item \textbf{Video Indexing:} Labeling faces in video.\cite{marcialis2003fusion}\\
\end{itemize}

\section{Conclusion}
Nowadays, face recognition is an unbounded exploration and development subject for research. We can use face recognition technology for remote employee tracking. Some extreme secure applications like person authentication or controlling entrance at certain areas are carried out using this technology. In our research, we try to represent a practical approach for recognition systems using the FaceNet model. Unlike the other existing methods, for example, principal component analysis (PCA) or support vector machine (SVM), our model doesn’t require any extra operations like classifying and grouping different images or creating a decision surface. Instead, our model shows improved performance on the recognition process as we used an end-to-end learning approach. Our method is more capable of improving the percentage of collecting valuable information and produce more distinct feature information, which basically increases the face recognition rate. \\
In our future works, we will try to improve our model by diminishing existing limitations and try to gain more efficiency. Again, a face spoofing attack is nowadays a big issue in security and authentication systems. So, we will make an attempt to develop an anti-spoofing method, which will help to improve the security of a biometric system.\\

%
%

\bibliographystyle{splncs03_unsrt}
\bibliography{ref}
\end{document}